\documentclass[runningheads]{llncs}
\usepackage[T1]{fontenc}
\usepackage{graphicx}
\usepackage{siunitx}
\usepackage{multirow}
\usepackage{makecell}
\usepackage{booktabs}
\usepackage{amsmath}
\usepackage{amssymb}
\usepackage{listings}
\usepackage{bbding}

\begin{document}

\title{SCREENER: A general framework for ta\underline{s}k-spe\underline{c}ific expe\underline{r}im\underline{e}nt d\underline{e}sig\underline{n} in quantitativ\underline{e} M\underline{R}I}
\titlerunning{SCREENER}
%
\author{Tianshu Zheng\inst{1 (}\Envelope\inst{)} \and Zican Wang\inst{2} \and Timothy Bray \inst{2} \and Daniel C. Alexander\inst{2} \and Dan Wu \inst{1 (}\Envelope\inst{)} \and Hui Zhang \inst{2} }

\authorrunning{Zheng et al.}

\institute{College of Biomedical Engineering \& Instrument Science, Zhejiang University,Hangzhou, Zhejiang, China, Hangzhou, China,\\
\and Centre for Medical Image Computing and Department of Computer Science, University College London, London, UK \\
\email{zhengtianshu@zju.edu.cn}\\
\email{danwu.bme@zju.edu.cn}
}

\maketitle              
\begin{abstract}
Quantitative magnetic resonance imaging (qMRI) is increasingly investigated for use in a variety of clinical tasks from diagnosis, through staging, to treatment monitoring. However, experiment design in qMRI, the identification of the optimal acquisition protocols, has been focused on obtaining the most precise parameter estimations, with no regard for the specific requirements of downstream tasks. Here we propose SCREENER: A general framework for ta\underline{s}k-spe\underline{c}ific expe\underline{r}im\underline{e}nt d\underline{e}sig\underline{n} in quantitativ\underline{e} M\underline{R}I. SCREENER incorporates a task-specific objective and seeks the optimal protocol with a deep-reinforcement-learning (DRL) based optimization strategy. To illustrate this framework, we employ a task of classifying the inflammation status of bone marrow using diffusion MRI data with intravoxel incoherent motion (IVIM) modelling. Results demonstrate SCREENER outperforms previous ad hoc and optimized protocols under clinical signal-to-noise ratio (SNR) conditions, achieving significant improvement, both in binary classification tasks, e.g. from 67\% to 89\%, and in a multi-class classification task, from 46\% to 59\%. Additionally, we show this improvement is robust to the SNR. Lastly, we demonstrate the advantage of DRL-based optimization strategy, enabling zero-shot discovery of near-optimal protocols for a range of SNRs not used in training. In conclusion, SCREENER has the potential to enable wider uptake of qMRI in the clinic.

\end{abstract}
\section{Introduction}
Conventional magnetic resonance imaging (cMRI) techniques are frequently used in clinical settings for the diagnosis and monitoring \cite{grover_magnetic_2015}. However, assessing subtle changes using cMRI can be challenging, because cMRI produces contrasts that reflect the aggregate effect from a host of tissue properties. To address this challenge, quantitative MRI (qMRI) has been developed to disentangle the specific contribution from each tissue property, enabling the estimation of each individually \cite{seiler_multiparametric_2021}. Owing to its potential to provide specific insights into tissue characteristics, qMRI has been increasingly investigated to improve diagnosis, staging and treatment monitoring \cite{keenan_recommendations_2019}. 

Despite its advantages over cMRI, qMRI has yet to achieve wide uptake clinically, due in significant part to its lengthier scan time requirement compared to cMRI, necessitated by the need to acquire data with a range of settings (e.g. different b-values in diffusion MRI (dMRI)). With the number of acquisition settings to include constrained by the scan time budget, it is crucial to optimize the experiment design so that only the most useful settings are included. As attempting this with an empirical, trial-and-error, approach, is time consuming, there has been significant interest in developing computational alternatives, with the most common approach being based on the Cramer-Rao lower bounds (CRLB) \cite{alexander_general_2008,boudreau_sensitivity_2018,cercignani2006optimal,lee_flexible_2019,leporq_optimization_2015,pena2020determination,poot2010optimal,wyatt2012comprehensive} which aims to minimize the parameter variability.

However, one common limitation of these existing approaches is that the underlying clinical task the qMRI experiment is designed for has no influence on the selection of the optimal protocols. We hypothesize that the optimal protocol may be task-dependent and propose a novel framework for qMRI experiment design that can be tailored for a particular task of interest. Our framework includes two components: a task-specific objective module and a deep reinforcement learning (DRL) based optimization module. The task-specific objective module leverages the recently developed, and validated, task-driven protocol assessment \cite{epstein_task-driven_2021}. The DRL-based optimization module is chosen over conventional optimization strategies because the former can “predict” near-optimal protocols under a different circumstance (e.g., different SNRs) to the one used during training without the need to repeat the optimization.

To evaluate our proposed framework, we employed a task of classifying the inflammation status of bone marrow using dMRI data with intravoxel incoherent motion (IVIM) modeling for four distinct classification tasks, including three binary classifications and one multi-class classification. Our main contributions can be summarized as follows:
\begin{enumerate}
    \item We propose a general computational framework for qMRI experiment design, which can optimize the protocol based on the specific task of interest.
    \item We show optimized protocols are task-dependent.
    \item We find optimized protocols can achieve significant, and robust, improvement in task performance over the ad hoc and CRLB-optimized protocols.
    \item We demonstrate DRL enables zero-shot discovery of near-optimal protocols across varying SNRs superior to ad hoc and CRLB-optimized protocols.
\end{enumerate}

The rest of the paper is organized as follows: Section 2 describes the methodology of our proposed SCREENER; Section 3 reports the evaluation strategy and details the results comparing the proposed method against the baselines; Sections 4 summarizes the contributions and discusses future work.

\section{Method}

Given a task of interest, SCREENER determines the optimal protocol through optimizing the corresponding task-specific objective. Below we first provide the details of the task-specific objective module before describing the specifics of the demonstrator of the proposed framework.

\subsection{Task-specific objective module}
\begin{figure}[t]
    \centering
    \includegraphics[height=0.25\textheight]{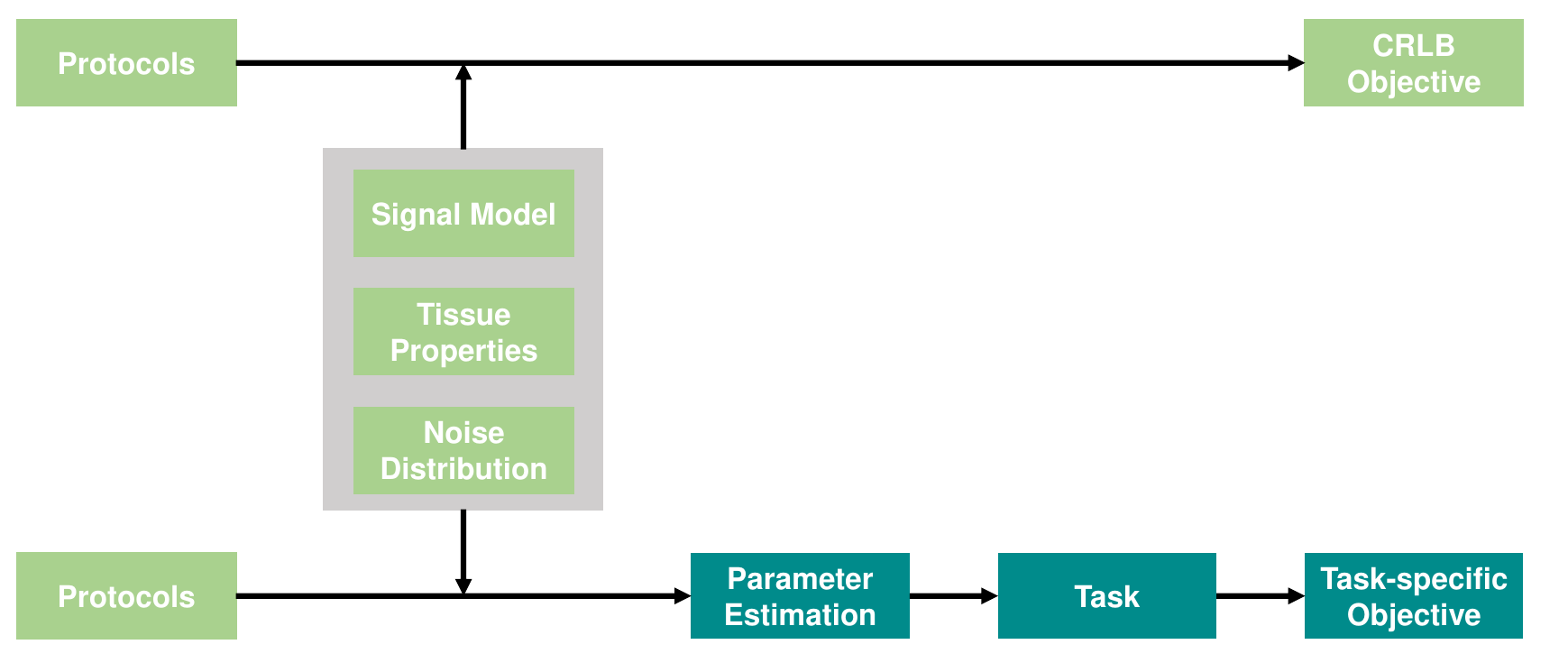}
    \caption{Our proposed task-specific objective function, SCREENER versus previous CRLB function. Components unique to SCREENER are colored in dark cyan.}
    \label{fig:fig1}
\end{figure}

The task-specific objective module adopts the algorithm recently described in~\cite{epstein_task-driven_2021}. Its overarching structure is depicted in Fig. \ref{fig:fig1}, alongside the popular CRLB module \cite{alexander_general_2008} for comparison. The differences between our proposed framework and previous CRLB work are indicated in dark cyan. 

The CRLB method is used to establish a lower limit for the variance of parameter estimates. However, it provides only a theoretical lower bound rather than an actual measure of parameter variability. Thus minimizing the CRLB does not guarantee a reduction in the actual variability of parameter estimation. But most importantly, it does not offer optimization tailored to a specific task, a problem we address here.

In contrast, the task-specific assessment addresses these limitations directly. It works by mimicking {\it in silico} real-world tasks (e.g. classification), downstream of parameter estimation, in its entirety. The algorithm can be summarised as follows:
\begin{enumerate}
    \item Simulate real-world cohorts with specific tissue properties;
    \item Simulate data acquisition from these cohorts;
    \item Determine tissue properties from the simulated data through parameter estimation;
    \item Evaluate task performance (e.g., classification) based on the estimated tissue properties.
\end{enumerate}

Common to both methods, a set of tissue properties to target must be specified prior to the optimization process. However, unique to the proposed method, tissue properties will be determined from a set of tissue types relevant to the task of interest. Each tissue type must have {\it a priori} known distribution of tissue properties. Tissue properties sampled from these known distributions are used to simulate the MRI signal.

Furthermore, in the proposed method, the generated signal is then passed to the parameter estimation module to estimate the respective tissue properties. These estimated tissue properties are then processed in accordance of the chosen task to obtain the task specific objective.


\subsection{Demonstration with IVIM}
In this study, we utilize a series of classification tasks to assess the effectiveness of our proposed framework. 
\\
\\
\noindent\textbf{Signal Model.} The dMRI data of IVIM modelling \cite{le_bihan_separation_1988} can be defined as:
\begin{equation}\label{eq:eq1}
S(b) = S_0 \, e^{\rm{-TE/T_2}} \left( f e^{-bD^\ast} + \left(1 - f\right) e^{-bD} \right)
\end{equation}
where $S(b)$ is the MRI signal at diffusion weighting b, $S_0$ is the signal at b = 0, TE is the echo time, T$_2$ is the relaxation time, $f$ is the perfusion fraction, $D^\ast$ is the pseudo-diffusivity, and $D$ is the diffusivity of non-perfusing water.

Unlike~\cite{epstein_task-driven_2021}, the IVIM model used here includes the dependence of T$_2$ relaxation. Our initial experiments using the version adopted in~\cite{epstein_task-driven_2021} suggest the CRLB approach prefers protocols with maximum b-values significantly lower than those preferred by the proposed framework (data not shown). As the maximum b-value dictates the minimum TE, thus the maximum SNR, it becomes necessary to account for the effect of TE through the modelling of T$_2$ relaxation.

The distribution of the parameters will follow Zhao et al. \cite{zhao_detection_2015} corresponding to the specific tissue types. After the noise free signals are generated, the Rician noise can be added as follows \cite{daducci_accelerated_2015}:

\begin{equation}\label{eq:eq2}
S_{noisy}(b)=\sqrt{{(S(b)+{\ \xi}_1)}^2+{\xi_2}^2}
\end{equation}
where $S_{noisy}(b)$ is the noise incorporated signal, ${\xi}_{1},{\xi}_{2}\sim \mathcal{N}(0,{\sigma}^2)$
, and $\sigma$=1/SNR.
\\
\\
\noindent\textbf{Parameter estimation.}
\setcounter{footnote}{0} %
We chose the most commonly used segmented fitting approach, building on Kaandrop et al. \cite{kaandorp_improved_2021}\footnote{https://github.com/oliverchampion/IVIMNET}. In this fitting process, we initially obtain the $D$ and $f$ from b-values $\geq 200 \rm{s/mm^2}$ \cite{zhao_detection_2015}. Subsequently, we estimate the value of $D^\ast$ from the residual signal. 
\\
\\
\noindent\textbf{Classification task.}
qMRI, especially the IVIM model, is a clinically important tool for characterizing bone marrow pathology \cite{dietrich2017diffusion}. One clinically-important example of this is in spondyloarthritis (SpA), an inflammatory disease affecting bones and joints, where this qMRI approach has been used previously \cite{zhao_detection_2015,wang2020comparative}. The existing literature provides estimates of particular tissue states which contribute to diagnosis and disease phenotyping in SpA: active, chronic, and healthy \cite{zhao_detection_2015}.

Distinguishing between these tissue types can be regarded either as a set of binary classification tasks or a single multi-class classification task. The binary classifications comprise three specific sub-tasks: active vs. chronic, active vs. healthy, and chronic vs. healthy. The multi-class classification involves distinguishing among active, chronic, and healthy classes.
\\
\\
\noindent\textbf{Classification model.}
Due to the lengthy computational requirements of parameter estimation and reinforcement learning, the k-nearest neighbors (KNN) algorithm is chosen for its efficiency \cite{cover_nearest_1967}. Unlike some methods, KNN does not necessitate an explicit training step; all computations take place during prediction. This classical non-parametric, instance-based learning method is widely utilized not only for binary but also for multi-class classification tasks. 

The input for the KNN model comprises four distinct parameters of the IVIM model: ${{[S}_0,f,D,D}^\ast]$, with the output being the accuracy of the prediction which we will use as the task-specific objective.
\\
\\
\noindent\textbf{DRL based optimization method.}
While any optimization algorithm may be used in principle, due to the expected high computational cost of the task-specific objective module, a DRL-based method is chosen, because it enables zero-shot prediction of near-optimal protocols. DRL learns the optimal b-value sampling policy for achieving the highest downstream task results.

In this study, we employ the popular proximal policy optimization (PPO) algorithm \cite{schulman_proximal_2017} which is known to be stable and robust. In particular, we use the implementation provided by stable-baseline3\footnote{https://github.com/DLR-RM/stable-baselines3}. DRL implementation uses the task-specific objective as the reward, with pre-specified set of tissue properties, qMRI signal model, SNR, and parameter estimation method forming the environment.
\\

\noindent\textbf{Implementation details.}
The original task-specific assessment in~\cite{epstein_task-driven_2021} was implemented in Matlab. To enable the integration of this assessment as a module with other Python-based tools, we have re-implemented this in Python for our study. We set the neighborhood of KNN to 5 and used 5-fold cross-validation to mitigate overfitting and the average accuracy is reported. Within the environment, 10 discrete actions, each corresponding to 1 b-value, have been individually defined, starting with an ad hoc protocol as the initial observation. In the PPO setting, both the actor and critic networks consist of 2 linear layers with 64 hidden units. The network was trained for 100,000 steps using the Adam optimizer with a learning rate of 1e-3. In the demonstrated task of classification, the classification accuracy is naturally chosen as reward. The policy gradient method we used is PPO.

\section{Experiments and Results}
This section consists of four parts: 1) description of dataset we used, 2) validation of the task-specific objective module, 3) demonstration of the framework's ability to identify optimal protocols that exhibit superior performance in three binary classification tasks and one multi-class classification scenario within a clinical context, and 4) robustness and zero-shot prediction assessment.

\subsection{Dataset}

In this study, MRI signals were simulated according to Eqns. [\ref{eq:eq1}-\ref{eq:eq2}]. To minimize the discrepancy between the simulation and real scanning, the IVIM parameter distributions for different tissue types were set to the values reported by Zhao et al. \cite{zhao_detection_2015}. We used a gradient strength (\textbf{G}) of 33 mT/m as reported in the official document\footnote{https://www.philips.com/refurbishedsystems} and T$_2$ was set as 100 ms following Wang et al., \cite{wang2020comparative}. The ad hoc protocol was also taken from Zhao et al. \cite{zhao_detection_2015} which employed a set of 10 b-values (0, 10, 20, 30, 50, 80, 100, 200, 400, and 800 $\rm{s/mm^2}$), and the SNR was set to 25 for $S_0$.


\subsection{Validation of the task-specific objective module}
In order to evaluate our Python implementation, we compared the performance for the three binary classification tasks, in terms of Area Under the Curve (AUC), with the original results from Zhao et al. \cite{zhao_detection_2015} and the results from the original Matlab implementation~\cite{epstein_task-driven_2021}. In these experiments, the populations of different groups follow Zhao et al., comprising an active group (n = 20), a chronic group (n = 21), and a healthy group (n = 21). The evaluation SNR follows Epstein et al. \cite{epstein_task-driven_2021}. We repeated the experiments 50 times and report the means and standard deviations of these fifty experiments, as shown in Table \ref{tab:tab1}. The results of the baselines are obtained directly from the corresponding papers.

In Table \ref{tab:tab1}, it can be seen that our Python implementation, is similar to the previous Matlab implementation, but exhibits higher similarity to the original results determined from clinical dataset. For example, in the classification of the active group and chronic group in the clinical dataset, only the parameter $D$ can be used for classification, while the other two parameters show similar performance (0.55 vs. 0.54). Our method can mimic this phenomenon (0.51 vs. 0.50) better than the original Matlab implementation (0.51 vs. 0.56) in terms of the mean AUC. This improvement may have resulted from the modelling of T$_2$ relaxation in our Python implementation.

\begin{table}[htbp]
\centering

\caption{Validation of the task-specific module: the AUCs estimated with our implementation is compared against those determined from clinical dataset~\cite{zhao_detection_2015} and those with the original implementation~\cite{epstein_task-driven_2021}.}
\label{tab:tab1}
\resizebox{\textwidth}{!}{%
\begin{tabular}{*{10}{c}}
\toprule
& \multicolumn{3}{c}{\textbf{Zhao et al.~\cite{zhao_detection_2015}}} & \multicolumn{3}{c}{\textbf{Epstein et al.~\cite{epstein_task-driven_2021}}} & \multicolumn{3}{c}{\textbf{Ours}} \\
\cmidrule(lr){2-4} \cmidrule(lr){5-7} \cmidrule(lr){8-10}
& \multicolumn{1}{p{1cm}}{\centering$f$} & \multicolumn{1}{p{1cm}}{\centering$D$} & \multicolumn{1}{p{1cm}}{\centering$D^*$} & \multicolumn{1}{p{1cm}}{\centering$f$} & \multicolumn{1}{p{1cm}}{\centering$D$} & \multicolumn{1}{p{1cm}}{\centering$D^*$} & \multicolumn{1}{p{1cm}}{\centering$f$} & \multicolumn{1}{p{1cm}}{\centering$D$} & \multicolumn{1}{p{1cm}}{\centering$D^*$} \\
\midrule
\textbf{Active} & \multirow{3}{*}{0.55} & \multirow{3}{*}{0.98} & \multirow{3}{*}{0.54} & \multirow{3}{*}{0.51} & \multirow{3}{*}{0.94} & \multirow{3}{*}{0.56} & \multirow{3}{*}{0.51$\pm$0.05} & \multirow{3}{*}{0.95$\pm$0.03} & \multirow{3}{*}{0.50$\pm$0.08} \\
\textbf{vs.} & & & & & & & & & \\
\textbf{Chronic} & & & & & & & & & \\
\hline
\textbf{Active} & \multirow{3}{*}{0.88} & \multirow{3}{*}{0.98} & \multirow{3}{*}{0.50} & \multirow{3}{*}{0.78} & \multirow{3}{*}{0.95} & \multirow{3}{*}{0.58} & \multirow{3}{*}{0.79$\pm$0.04} & \multirow{3}{*}{0.96$\pm$0.03} & \multirow{3}{*}{0.52$\pm$0.08} \\
\textbf{vs.} & & & & & & & & & \\
\textbf{Healthy} & & & & & & & & & \\
\hline
\textbf{Chronic} & \multirow{3}{*}{0.94} & \multirow{3}{*}{0.50} & \multirow{3}{*}{0.53} & \multirow{3}{*}{0.84} & \multirow{3}{*}{0.53} & \multirow{3}{*}{0.65} & \multirow{3}{*}{0.84$\pm$0.04} & \multirow{3}{*}{0.53$\pm$0.03} & \multirow{3}{*}{0.50$\pm$0.07} \\
\textbf{vs.} & & & & & & & & & \\
\textbf{Healthy} & & & & & & & & & \\
\bottomrule
\end{tabular}%
}
\end{table}

\subsection{Performance of the overall framework}
After validating the correctness of the task-specific objective module, we evaluate the ability of the proposed framework to identify task-specific optimal protocols on three binary classification tasks and a multi-class classification task. We compare our optimized protocols with the ad hoc protocol reported in Zhao et al. \cite{zhao_detection_2015} and a standard protocol optimization baseline using CRLB \cite{alexander_general_2008}. 
Both the CRLB optimization framework and our proposed method are optimized under the same SNR and tissue types. For the classification results, we conduct five-fold cross-validation, with the same population groups as Zhao et al. \cite{zhao_detection_2015}. We repeat the classification process 50 times and report the mean accuracy along with the standard deviation.
\\
\\
\noindent\textbf{Binary classification.}
The results of the optimized protocols and the classification accuracies are shown in Table \ref{tab:tab2}. It can be seen that among all three tasks, the ad hoc protocol shows the worst accuracies, the CRLB method shows moderate results, and our proposed method demonstrates the highest accuracy with at least a 3\% improvement compared to the ad hoc sequence and a 2\% improvement over the CRLB method. Furthermore, for the active group and chronic group, which are of greater clinical interest, our method achieved an improvement of up to 10\% compared to the previous optimization method (CRLB).

\begin{table}[htbp]
\centering
\caption{The optimized protocols and accuracies of the three binary classification tasks with the Ad hoc method, CRLB method, and SCREENER method.}
\label{tab:tab2}
\resizebox{\textwidth}{!}{%
\begin{tabular}{ccccccc} 
\toprule
& \multicolumn{2}{c}{\textbf{Ad hoc}} & \multicolumn{2}{c}{\textbf{CRLB}} & \multicolumn{2}{c}{\textbf{SCREENER}} \\ 
\cmidrule(lr){2-3} \cmidrule(lr){4-5} \cmidrule(l){6-7}
& \textbf{Sequence} & \textbf{Accuracy} & \textbf{Sequence} & \textbf{Accuracy} & \textbf{Sequence} & \textbf{Accuracy} \\ 
\midrule
{\textbf{Active}} & &  & \multirow{1}{*}{\begin{tabular}[c]{@{}c@{}}b=0,0,7,\\7,7,7,52\\52,52,508\end{tabular}} &  & \multirow{1}{*}{\begin{tabular}[c]{@{}c@{}}b=0,212,254\\272,356,530\\600,731,929,959\end{tabular}} &  \\
\textbf{vs.} & & \multirow{1}{*}{0.66$\pm$0.06} & & \multirow{1}{*}{0.74$\pm$0.05} & & \multirow{1}{*}{0.87$\pm$0.04} \\
\textbf{Chronic} & & & & & & \\
\cmidrule{1-1} \cmidrule{3-3} \cmidrule{4-5} \cmidrule{6-7}
{\textbf{Active}} & \multirow{1}{*}{\begin{tabular}[c]{@{}c@{}}b=0,10,20,\\30,50,80,\\100,200,400,800\end{tabular}} &  & \multirow{1}{*}{\begin{tabular}[c]{@{}c@{}}b=0,0,7,\\7,7,7,52,\\52,52,470\end{tabular}} &  & \multirow{1}{*}{\begin{tabular}[c]{@{}c@{}}b=0,221,435\\483,570,667\\750,765,893,936\end{tabular}} &  \\
\textbf{vs.} & & \multirow{1}{*}{0.67$\pm$0.06} & & \multirow{1}{*}{0.74$\pm$0.04} & & \multirow{1}{*}{0.89$\pm$0.04}\\
\textbf{Healthy} & & & & & & \\
\cmidrule{1-1} \cmidrule{3-3} \cmidrule{4-5} \cmidrule{6-7}
{\textbf{Chronic}} &  &  & \multirow{1}{*}{\begin{tabular}[c]{@{}c@{}}b=0,0,7,\\7,7,7,52,\\52,52,478\end{tabular}} &  & \multirow{1}{*}{\begin{tabular}[c]{@{}c@{}}b=0,0,143,\\329,364,551,\\631,635,664,784\end{tabular}}  \\
\textbf{vs.} & & \multirow{1}{*}{0.51$\pm$0.07} & & \multirow{1}{*}{0.52$\pm$0.06} &  & \multirow{1}{*}{0.54$\pm$0.06}\\

\textbf{Healthy} & & & & & & \\
\bottomrule
\end{tabular}
}
\end{table}

\noindent\textbf{Multi-class classification.}
We further test our proposed method on the multi-class classification task, which is representative of the real-world clinical scenario where multiple tissue types are possible and need to be distinguished. Similar results to those seen with the binary classification tasks are observed: ad hoc has the worst accuracy and SCREENER has the highest accuracy. Our proposed method achieves a 13\% improvement compared to the ad hoc protocol and 11\% compared to the CRLB-optimized protocol. Additionally, the CRLB-optimized protocol for this multi-class classification is identical to the CRLB-optimized protocol for the binary classification task (chronic vs. healthy), which underscores the non-task-specific nature of the CRLB approach.
\begin{table}[htbp]
\centering
\caption{The optimized protocols and accuracies of the multi-class classification task with the Ad hoc method, CRLB method, and SCREENER method}
\label{tab:tab3}
\begin{tabular}{@{}lccc@{}}
\toprule
\textbf{Method}                    & \textbf{Ad hoc}       & \textbf{CRLB}  & \textbf{SCREENER}           \\ 
\midrule
\multirow{2}{*}{\textbf{Sequence}} & b=$0,10,20,30,50,80,$ & b=$0,0,7,7,7,$ & b=$0,175,229,336,540$        \\
                                   & $100,200,400,800$     & $7,52,52,52,478$    & $595,603,618,629,881$  \\ 
\hline
\textbf{Accuracy}                  & $0.46\pm0.05$         & $0.48\pm0.05$  & $0.59\pm0.04$               \\
\bottomrule
\end{tabular}
\end{table}

\subsection{Robustness and the zero-shot ability of the framework}
\begin{figure}
    \centering
    \includegraphics[width=1\linewidth]{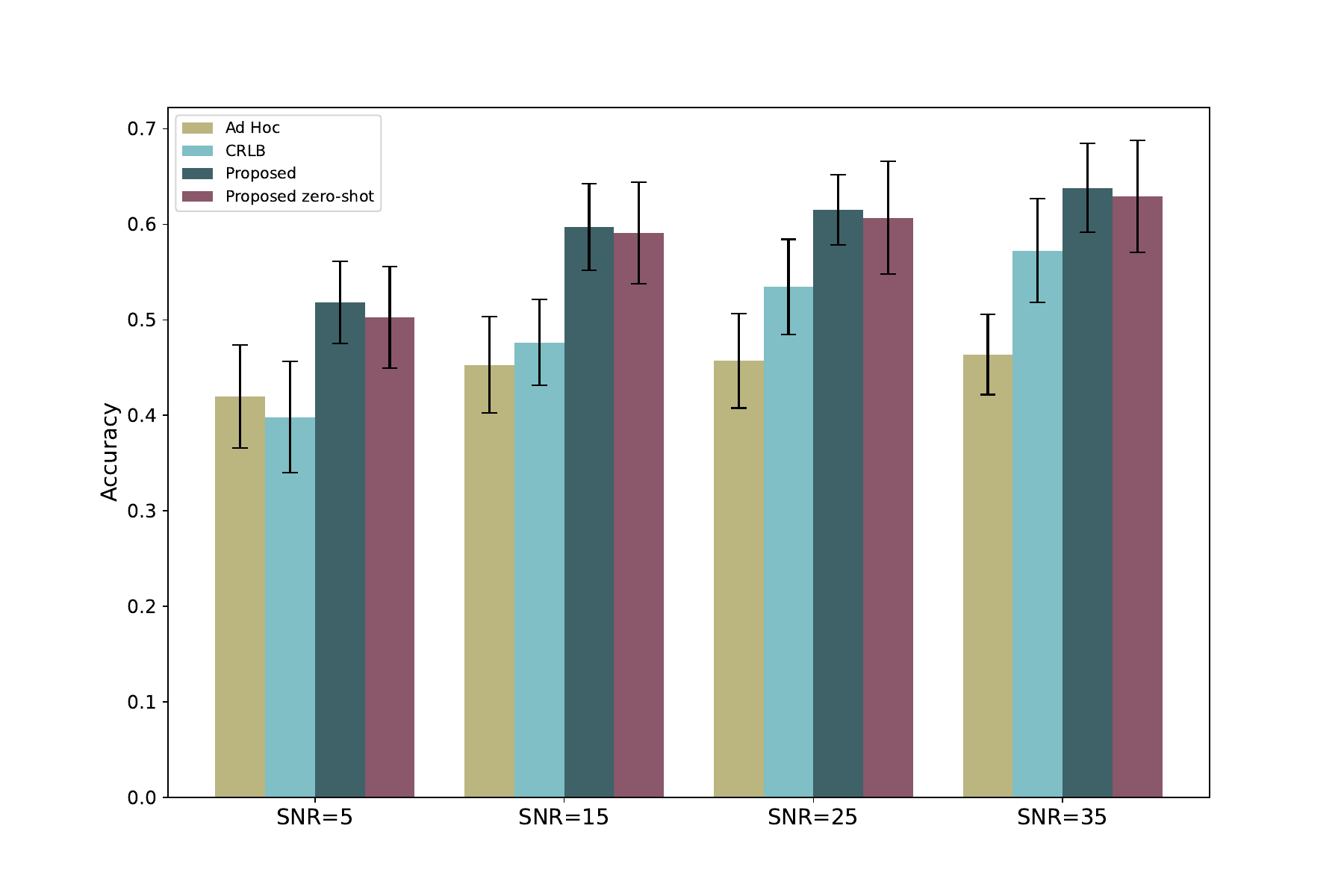}
    \caption{The accuracies of the multi-class classification task with the Ad hoc method, CRLB method, and SCREENER, SCREENER (zero-shot) under different SNR.}
    \label{fig:fig2}
\end{figure}

To thoroughly assess the effectiveness of our proposed framework, we conducted robustness and zero-shot prediction tests on the multi-class classification task. 
All the settings were the same as in Section 3.3 for multi-class classification, except for the SNR. The results can be found in Fig. \ref{fig:fig2}. 
\\
\\
\noindent\textbf{Varied SNR.} We trained the CRLB and the SCREENER under varied SNR and tested them under the same conditions to show the robustness of the SCREENER. The results are consistent with the previous experiments in Section 3.3. Our proposed method achieves the highest accuracy, while the ad hoc method achieves the lowest. Furthermore, we observe that both optimized protocols (CRLB and SCREENER) achieve higher accuracy as SNR increases. In contrast, the ad hoc protocol exhibits lower accuracy with increasing SNR.
\\
\\
\noindent\textbf{Zero-shot SNR.} As it can be practically difficult to estimate SNR in clinical protocols, we also tested the zero-shot ability of our method against training with incorrect SNR. Specifically, we trained our method on clinical SNR=25 and tested it on datasets with SNR=[5, 15, 25, 35]. As depicted in Fig. \ref{fig:fig2}, our method achieves the highest accuracy compared with other methods, which exhibits zero-shot discovery ability of near-optimal protocols. Additionally, our method shows similar trends to those seen in the Varied SNR tests: higher SNR correlates with higher accuracy.

\section{Conclusion}
In this paper, we present a novel framework for task-specific experiment design optimization in qMRI and demonstrate its impact on a clinically-relevant classification task. Our work showcases that the optimized protocols are task-dependent and can lead to significant improvements compared to previous approaches. Additionally, we demonstrate the robustness of the framework across various levels of SNRs and its ability to discover near-optimal protocols in zero-shot scenarios. In the future, we plan to test the framework on additional models and tasks to evaluate its generalizability, such as the time-dependent dMRI model for glioma classification. \cite{zhang2023histological}.

{
\bibliographystyle{splncs04}
\bibliography{reference.bib}

\begin{thebibliography}{10}
\providecommand{\url}[1]{\texttt{#1}}
\providecommand{\urlprefix}{URL }
\providecommand{\doi}[1]{https://doi.org/#1}

\bibitem{alexander_general_2008}
Alexander, D.C.: A general framework for experiment design in diffusion {MRI} and its application in measuring direct tissue-microstructure features. Magnetic Resonance in Medicine: An Official Journal of the International Society for Magnetic Resonance in Medicine  \textbf{60}(2),  439--448 (2008)

\bibitem{boudreau_sensitivity_2018}
Boudreau, M., Pike, G.B.: Sensitivity regularization of the {Cramér}-{Rao} lower bound to minimize {B1} nonuniformity effects in quantitative magnetization transfer imaging. Magnetic Resonance in Medicine  \textbf{80}(6),  2560--2572 (2018)

\bibitem{cercignani2006optimal}
Cercignani, M., Alexander, D.C.: Optimal acquisition schemes for in vivo quantitative magnetization transfer mri. Magnetic Resonance in Medicine: An Official Journal of the International Society for Magnetic Resonance in Medicine  \textbf{56}(4),  803--810 (2006)

\bibitem{cover_nearest_1967}
Cover, T., Hart, P.: Nearest neighbor pattern classification. IEEE transactions on information theory  \textbf{13}(1),  21--27 (1967)

\bibitem{daducci_accelerated_2015}
Daducci, A., Canales-Rodríguez, E.J., Zhang, H., Dyrby, T.B., Alexander, D.C., Thiran, J.P.: Accelerated {Microstructure} {Imaging} via {Convex} {Optimization} ({AMICO}) from diffusion {MRI} data. NeuroImage  \textbf{105},  32--44 (2015)

\bibitem{dietrich2017diffusion}
Dietrich, O., Geith, T., Reiser, M.F., Baur-Melnyk, A.: Diffusion imaging of the vertebral bone marrow. NMR in Biomedicine  \textbf{30}(3),  e3333 (2017)

\bibitem{epstein_task-driven_2021}
Epstein, S.C., Bray, T.J., Hall-Craggs, M.A., Zhang, H.: Task-driven assessment of experimental designs in diffusion {MRI}: {A} computational framework. Plos one  \textbf{16}(10),  e0258442 (2021)

\bibitem{grover_magnetic_2015}
Grover, V.P., Tognarelli, J.M., Crossey, M.M., Cox, I.J., Taylor-Robinson, S.D., McPhail, M.J.: Magnetic {Resonance} {Imaging}: {Principles} and {Techniques}: {Lessons} for {Clinicians}. J Clin Exp Hepatol  \textbf{5}(3),  246--255 (Sep 2015), \url{https://www.ncbi.nlm.nih.gov/pmc/articles/PMC4632105/}

\bibitem{kaandorp_improved_2021}
Kaandorp, M.P., Barbieri, S., Klaassen, R., van Laarhoven, H.W., Crezee, H., While, P.T., Nederveen, A.J., Gurney-Champion, O.J.: Improved unsupervised physics-informed deep learning for intravoxel incoherent motion modeling and evaluation in pancreatic cancer patients. Magnetic Resonance in Medicine  \textbf{86}(4),  2250--2265 (2021)

\bibitem{keenan_recommendations_2019}
Keenan, K.E., Biller, J.R., Delfino, J.G., Boss, M.A., Does, M.D., Evelhoch, J.L., Griswold, M.A., Gunter, J.L., Hinks, R.S., Hoffman, S.W., {others}: Recommendations towards standards for quantitative {MRI} ({qMRI}) and outstanding needs. Journal of magnetic resonance imaging: JMRI  \textbf{49}(7), ~e26 (2019)

\bibitem{le_bihan_separation_1988}
Le~Bihan, D., Breton, E., Lallemand, D., Aubin, M.L., Vignaud, J., Laval-Jeantet, M.: Separation of diffusion and perfusion in intravoxel incoherent motion {MR} imaging. Radiology  \textbf{168}(2),  497--505 (Aug 1988)

\bibitem{lee_flexible_2019}
Lee, P.K., Watkins, L.E., Anderson, T.I., Buonincontri, G., Hargreaves, B.A.: Flexible and efficient optimization of quantitative sequences using automatic differentiation of {Bloch} simulations. Magnetic resonance in medicine  \textbf{82}(4),  1438--1451 (2019)

\bibitem{leporq_optimization_2015}
Leporq, B., Saint-Jalmes, H., Rabrait, C., Pilleul, F., Guillaud, O., Dumortier, J., Scoazec, J.Y., Beuf, O.: Optimization of intra-voxel incoherent motion imaging at 3.0 {Tesla} for fast liver examination. Journal of Magnetic Resonance Imaging  \textbf{41}(5),  1209--1217 (2015)

\bibitem{pena2020determination}
Pena-Nogales, O., Hernando, D., Aja-Fernandez, S., de~Luis-Garcia, R.: Determination of optimized set of b-values for apparent diffusion coefficient mapping in liver diffusion-weighted mri. Journal of Magnetic Resonance  \textbf{310},  106634 (2020)

\bibitem{poot2010optimal}
Poot, D.H., den Dekker, A.J., Achten, E., Verhoye, M., Sijbers, J.: Optimal experimental design for diffusion kurtosis imaging. IEEE transactions on medical imaging  \textbf{29}(3),  819--829 (2010)

\bibitem{schulman_proximal_2017}
Schulman, J., Wolski, F., Dhariwal, P., Radford, A., Klimov, O.: Proximal policy optimization algorithms. arXiv preprint arXiv:1707.06347  (2017)

\bibitem{seiler_multiparametric_2021}
Seiler, A., Nöth, U., Hok, P., Reiländer, A., Maiworm, M., Baudrexel, S., Meuth, S., Rosenow, F., Steinmetz, H., Wagner, M., {others}: Multiparametric quantitative {MRI} in neurological diseases. Frontiers in Neurology  \textbf{12},  640239 (2021)

\bibitem{wang2020comparative}
Wang, D., Yin, H., Liu, W., Li, Z., Ren, J., Wang, K., Han, D.: Comparative analysis of the diagnostic values of t2 mapping and diffusion-weighted imaging for sacroiliitis in ankylosing spondylitis. Skeletal Radiology  \textbf{49},  1597--1606 (2020)

\bibitem{wyatt2012comprehensive}
Wyatt, C., Soher, B.J., Arunachalam, K., MacFall, J.: Comprehensive analysis of the cramer--rao bounds for magnetic resonance temperature change measurement in fat--water voxels using multi-echo imaging. Magnetic Resonance Materials in Physics, Biology and Medicine  \textbf{25},  49--61 (2012)

\bibitem{zhang2023histological}
Zhang, H., Liu, K., Ba, R., Zhang, Z., Zhang, Y., Chen, Y., Gu, W., Shen, Z., Shu, Q., Fu, J., et~al.: Histological and molecular classifications of pediatric glioma with time-dependent diffusion mri-based microstructural mapping. Neuro-oncology  \textbf{25}(6),  1146--1156 (2023)

\bibitem{zhao_detection_2015}
Zhao, Y.h., Li, S.l., Liu, Z.y., Chen, X., Zhao, X.c., Hu, S.y., Liu, Z.h., Ms, Y.j.M., Chan, Q., Liang, C.h.: Detection of active sacroiliitis with ankylosing spondylitis through intravoxel incoherent motion diffusion-weighted {MR} imaging. European radiology  \textbf{25},  2754--2763 (2015)

\end{thebibliography}
}

\end{document}